\documentclass[runningheads]{llncs}

\usepackage{graphicx}

\usepackage{subcaption}
\usepackage{hyperref}

\usepackage{multirow}
\usepackage{amsmath}
\usepackage{float}

\usepackage{algorithm}
\usepackage{algpseudocode}
\usepackage{algorithmicx}  
\usepackage{varwidth}

\usepackage[usenames, dvipsnames]{color}

\begin{document}

\title{Deep Learning Based Vehicle Make-Model Classification}

\author{Burak Satar\inst{1} \and
Ahmet Emir Dirik\inst{2}\thanks{Corresponding Author}}

\institute{Uludag University, Bursa, Turkey\\
    Department of Electrical-Electronics Engineering\\
    \email{buraksatar@gmail.com} 
\and
    Uludag University, Bursa, Turkey\\
    Department of Computer Engineering\\
    \email{edirik@uludag.edu.tr}
}

\maketitle

\begin{abstract}
This paper studies the problems of vehicle make \& model classification. Some of the main challenges are reaching high classification accuracy and reducing the annotation time of the images. To address these problems, we have created a fine-grained database using online vehicle marketplaces of Turkey. A pipeline is proposed to combine an SSD (Single Shot Multibox Detector) model with a CNN (Convolutional Neural Network) model to train on the database. In the pipeline, we first detect the vehicles by following an algorithm which reduces the time for annotation. Then, we feed them into the CNN model. It is reached approximately 4\% better classification accuracy result than using a conventional CNN model. Next, we propose to use the detected vehicles as ground truth bounding box (GTBB) of the images and feed them into an SSD model in another pipeline. At this stage, it is reached reasonable classification accuracy result without using perfectly shaped GTBB. Lastly, an application is implemented in a use case by using our proposed pipelines. It detects the unauthorized vehicles by comparing their license plate numbers and make \& models. It is assumed that license plates are readable.

\keywords{deep learning \and vehicle \and model \and classification \and CNN \and ResNet \and detection \and SSD \and fraud \and license plate}
\end{abstract}

\section{Introduction}

Numerous researches have been performed on make \& model classification of vehicles \cite{vmm1,vmm2,vmm3}. This paper studies some of the main issues of these researches. First of all, there are only a few open source databases which include various vehicles. However, they generally either contain fewer images per class or don't include commonly used vehicles. In this study, a country-specific database is created to address this issue. Table 1 and 2 explain its content. Images are collected through online sources such as vehicle marketplace websites. A script is used as a web crawler to gather the images.

On the other hand, plenty of related studies use only CNN \cite{convolution} based architectures. Thus, they can't reach high classification accuracy results. We implement three different experiments in our database to approach this problem. As a first pipeline, normalization and data-augmentation processes are applied to the database. Then, we feed the images into a ResNet (Residual Network) \cite{resnet} model for classification. This is called Experiment I. As a second pipeline, the vehicles are detected by an SSD \cite{ssd} model which pre-trained on MS COCO \cite{coco} and PASCAL VOC \cite{pascal} databases. Detected vehicles pass through the same pre-processing methods. They are fed into the same ResNet model for classification. We call it Experiment II. It is shown that Experiment II reaches a higher classification accuracy result than Experiment I.

Moreover, annotation of the GTBB is another issue. Using manual annotation tools \cite{annotation_manual} requires a considerable amount of time when a database is immense. For instance, three million images should be annotated manually in some studies \cite{makemodel_Dehghan}. Using methods like Amazon Mechanical Turk also would be possible; however, it could cost a lot of money when the database contains a relatively high volume of the data. For those reasons, we use the pre-trained SSD model for annotation which we already use in Experiment II. Algorithm 1 explains how we implement the annotation semi-autonomously in our database. Therefore, the coordinates of detected vehicles are picked as GTBB of the images. We fine-tune the VGG \cite{vgg} based SSD model on our database. This pipeline is called Experiment III. It is reached a relatively close classification accuracy result when comparing to Experiment I. It is seen that Experiment III achieves this score without having perfectly shaped GTBB. 

Besides, we propose an application to implement this study in a use case. The use case is regarding the detection of an illegal vehicle. Plenty of studies handle detecting the license plates \cite{alpr,licensePlate}. However, it is challenging to understand which vehicle uses a fraudulent license plate when a recurrent license plate is recognized. For this reason, we suggest using our vehicle make \& model classification methods to detect the unauthorized license plates assuming that license plates are readable. Reading the license plates is out of the scope of this study. Thus, an open source project is used to fulfill the need \cite{openSourcelicensePlate}.

The remainder of this paper is organized as follows. Section 2 provides all the details of our system: data gathering, annotation, testing models, classification and detection architectures respectively.  Section 3 describes the experiments with results. Section 4 presents the conclusions and future studies.

\section{Vehicle Make-Model Classification}

Firstly, this part explains the details about the database. Then, it introduces an algorithm to annotate the database for detection purposes. Lastly, the components of the testing models are explained. 

\subsection{Data Gathering}

We do all experiments in the database; therefore it has a crucial effect on the results. Table 1 shows the distribution of the database. We take into account the statistical works of TurkStat (Turkish Statistical Institute) \cite{url_dataset} to form the database. The Institute monthly declares the number of vehicle brands which are registered in Turkey. According to the top 5 list which is composed of between 2015 and 2017, we choose Volkswagen since it is at the top of the list. We choose the models of Renault and Fiat because they manufacture local models in our town. They are also on top 5 of the list. It is also needed to indicate that Fiat Dogan SLX and Renault R12 Toros are manufactured in Turkey and only sold in Turkey. For the seventh class which stands for make \& models of other vehicle brands other than Volkswagen, Renault, Fiat; the statistical data of TurkStat is also taken into account. Therefore, this work becomes more focused on country-specific data.

Several focused keywords are used to gather the classes of the images. We eliminate the ones that have inappropriate features such as showing the inside of the vehicle, containing not the main part of the vehicle, etc. As a result, the database includes 27887 number of images of the vehicles.  

\begin{table}[H]
\centering
\caption{Distribution of the dataset}
\label{my-label}
\begin{tabular}{|l|c|c|c|c|}
\hline
\multicolumn{1}{|c|}{Make}       & Model   & Year & Feature                        & \# of Images \\ \hline
Volkswagen & Passat  & 2015 & 1.6 TDi BlueMotion Comfortline & 4024         \\ \hline
Renault    & Fluence & 2016 & 1.5 dCi Touch                  & 4293         \\ \hline
Fiat       & Linea   & 2013 & 1.3 Multijet Active Plus       & 4234         \\ \hline
Volkswagen & Polo    & 1999 & 1.6                            & 3208         \\ \hline
Renault    & Toros   & 2000 & R12                            & 3783         \\ \hline
Fiat       & Dogan   & 1996 & SLX                            & 4183         \\ \hline
\multicolumn{4}{|c|}{Other Class}                                 & 4162         \\ \hline
\end{tabular}
\end{table}

Table 2 shows the distribution of the seventh class which refers to the other cars in general. It is composed of seven different make \& models, other than the first six classes. 

\begin{table}[H]
\centering
\caption{Distribution of the other class}
\label{my-label}
\begin{tabular}{|l|c|c|c|c|}
\hline
\multicolumn{1}{|c|}{Make} & Model    & Year & Feature          & \# of Images \\ \hline
Toyota                     & Corolla  & 2016 & 1.4 D-4D Advance & 663          \\ \hline
Volvo                      & S60      & 2014 & 1.6 D Premium    & 707          \\ \hline
Peugeot                    & 206      & 2001 & 1.4 XR           & 468          \\ \hline
Ford                       & Focus    & 2017 & 1.6 TDCi Trend X & 693          \\ \hline
Mercedes-Benz              & C        & 2015 & CLA 180d         & 608          \\ \hline
Nissan                     & Micra    & 2016 & 1.2 Match        & 533          \\ \hline
Audi                       & A3 Sedan & 2017 & 1.6 TDI          & 490          \\ \hline
\end{tabular}
\end{table}

\subsection{Annotation}

We follow Algorithm 1 to reduce the annotation time. It defines the GTBB and classes of the images from the database with the help of predicted outcomes of pre-trained SSD model. In this case, we assume that the images usually include a car which is bigger than a certain size. Annotation takes a work day long when this algorithm is implemented in our database.
\vspace{-0.3cm}
\noindent
\begin{algorithm}[H]
  \caption{Annotating ground truth bounding boxes and classes  
  \label{alg:annotation}}  
  \begin{algorithmic}  
    \Require{certainSize $\gets$ a threshold value, classId $\gets$ zero,}  
    \For{all images of that class}
        \State{read the image and pass it through the pre-trained SSD to detect only cars;}
        \State{carSize $\gets$ the size of the detected car};
        \If{carSize $\geq$ certainSize} \par
          \hspace{\algorithmicindent} classOfDetectedCar $\gets$ classId;
        \Else \par
          \hspace{\algorithmicindent} ask annotator to give a label or delete it;
        \EndIf \par
    save the annotation to a .csv file;
    \EndFor \par 
\noindent increase the classID by one if any and run again;
\end{algorithmic}
\end{algorithm}
\noindent 
\vspace{-1cm}

\subsection{Model Training and Testing}
\noindent
\begin{figure}[H]
\begin{center}
\includegraphics[width=0.99\linewidth]{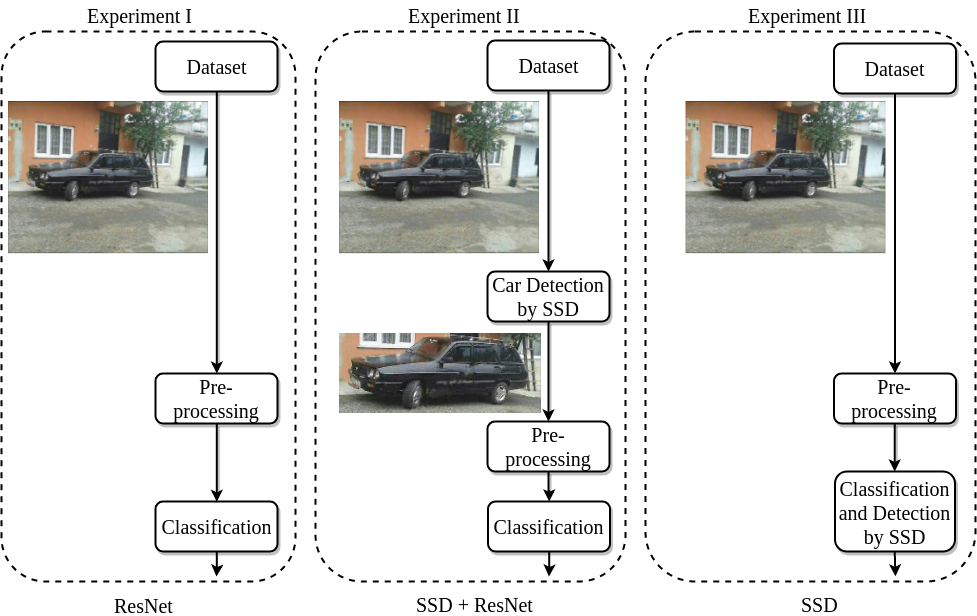}
\end{center}
\caption{Overview of experiments}
\end{figure}

Figure 1 presents the three main experiments. It shows the differences among classification results of using a custom ResNet model only, a pre-trained SSD with the ResNet model and a fine-tuned SSD only. 

Figure 2 shows the custom designed ResNet model which has 30 layers. It is used in Experiment I and II for classification by one difference. In Experiment I, the model takes the images to implement the pre-processing methods on them. Pre-processing methods include normalization, zero padding, resizing and data augmentation. The images are resized to the shape of (300,300,3). Data augmentation is done by flipping, adding Gaussian blur, adding Gaussian noise and zooming. Later, they are fed into the model for training. In Experiment II images are processed through an SSD model, which has weights pre-trained on MS COCO and fine-tuned on PASCAL VOC07 \& VOC12 database, to only detect cars. Then, the same pre-processing methods are applied to the images of the detected vehicles. They are fed into the same ResNet model for training. Therefore we give only the vehicles to the model in Experiment II instead of giving the whole image to the model.

The coordinates of the vehicles are detected in Experiment II. They are used as ground truth bounding boxes of the database in Experiment III. We take VGG based weights for the SDD model which are pre-trained on ImageNet. Then, the SSD model is fine-tuned on our database for classification and detection purposes.

\begin{gather}
 y_{predict}^{I,II} =
  \begin{bmatrix}
   [class_1 : prob_1], 
   & 
   [class_2 \hspace{0.1cm}:\hspace{0.1cm} prob_2], 
   & ..., 
   & 
   [class_7 : prob_7]
   \end{bmatrix}
\end{gather}
\begin{gather}
 y_{predict}^{III} =
  \begin{bmatrix}
   [prob, & class, & x_{min}, & y_{min}, & x_{max}, & y_{max}], & [...] 
   \end{bmatrix}
\end{gather}
Equation 1 refers to the outcome of Experiment I and II. However, Equation 2 refers to the output of Experiment III. Probability can be in a range between 0 and 1. 

\begin{figure}[H]
\resizebox{\textwidth}{!}{%
\begin{subfigure}{0.5\textwidth}
    \includegraphics[width=0.95\linewidth]{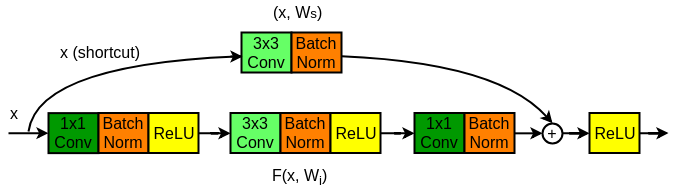}
    \caption{Convolutional Block}
    \label{fig:subim10}
    \vspace{5mm}
    \end{subfigure}
    \begin{subfigure}{0.5\textwidth}
    \includegraphics[width=0.95\linewidth]{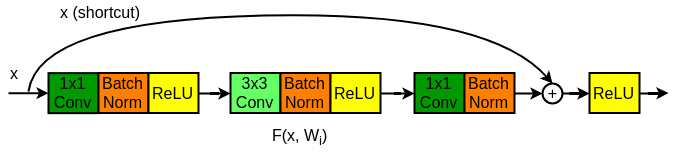}
    \caption{Identity Block}
    \label{fig:subim11}
    \vspace{2mm}
\end{subfigure}
}
\begin{subfigure}{1\textwidth}
    \includegraphics[width=1\linewidth]{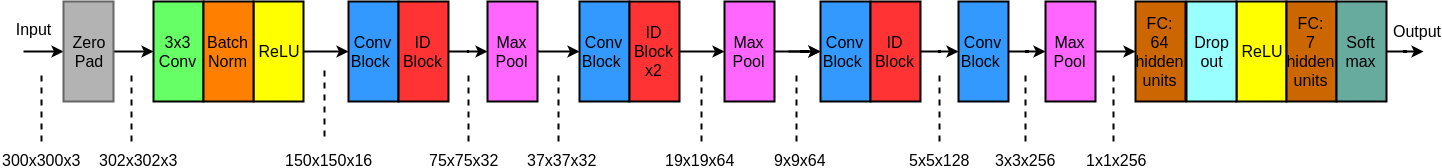}
    \caption{Main Block}
    \label{fig:subim12}
\end{subfigure}
\caption{The architecture of the ResNet based classification model}
\label{fig:image10}
\end{figure}

The number of trainable parameters which we use in the architecture is equal to 1,132,775. (1,1) is used as stride values in convolutional sections of Identical Blocks. Thus, heights and widths keep their shapes the same. The filter sizes are also not changed. Equation 3 shows the output of the block.

In Convolutional Block, the first section and shortcut section have a stride of (2,2) while other sections have a stride of (1,1). Besides, the first and second section of the main branch has equal filter number. However, the third section on Main Branch has same filter number with the shortcut section. Equation 4 shows the output of the block.

\begin{equation}
  y_{identity\_block} = \textit{F}(x, {W_i}) + x \;  .
\end{equation}
\begin{equation}
  y_{conv\_block} = \textit{F}(x, {W_i}) + W_sx \;  .
\end{equation}

Figure 3 shows the architecture of the SSD model. It is consist of series of convolutional blocks. Detections are made on certain levels. The number of detections is equal to 8732 per class. Then, a Non-Maximum Suppression method is implemented to eliminate predictions that have low Intersection over Union (IoU) ratio with GTBB. 
\begin{figure}[H]
\begin{center}
\includegraphics[width=1\linewidth]{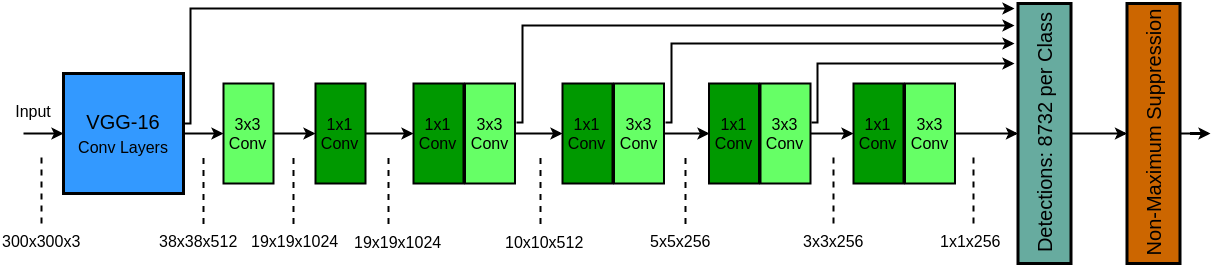}
\end{center}
\caption{The architecture of the SSD model for detection}
\end{figure}

\section{Experimental Results}

In this part, we examine the results with the help of Table 3, confusion matrices, sample outcomes. 
Training-validation-test sets are distributed based on 80\%-10\%-10\% rule in Experiment I and II. We implement the distribution of 80\%-20\% on training-test sets in Experiment III. NVIDIA GT 730 with 2GB RAM is used for Experiment I and II. However, Tesla K80 is used with the help of Google Colab for Experiment III because it needs a lot more computation. 

\begin{table}[]
\centering
\caption{Comparing the results of experiments}
\vspace{0.4cm}
\label{my-label}
\begin{tabular}{c|c|c|c|}
\cline{2-4}
                                  & Experiment I & Experiment II                                                       & Experiment III                                                        \\ \hline
\multicolumn{1}{|c|}{Method}      & Only ResNet  & \begin{tabular}[c]{@{}c@{}}Pre-trained SSD + \\ ResNet\end{tabular} & \begin{tabular}[c]{@{}c@{}}SSD with \\ VGG based Weights\end{tabular} \\ \hline
\multicolumn{1}{|c|}{Batch Size}  & \multicolumn{2}{c|}{32}                                                            & 32                                                                    \\ \hline
\multicolumn{1}{|c|}{Epoch}       & \multicolumn{2}{c|}{100}                                                           & 30                                                                    \\ \hline
\multicolumn{1}{|c|}{Loss}        & \multicolumn{2}{c|}{Categorical Cross Entropy}                                     & Smooth L1 + Softmax                                                   \\ \hline
\multicolumn{1}{|c|}{Train score} & 0.9635       & 0.9836                                                              & 0.9071                                                                \\ \hline
\multicolumn{1}{|c|}{Valid score} & 0.9052       & 0.9376                                                              & \multirow{2}{*}{0.9057}                                               \\ \cline{1-3}
\multicolumn{1}{|c|}{Test score}  & 0.9127       & 0.9510                                                              &                                                                       \\ \hline
\end{tabular}
\end{table}

Table 3 and Figure 4 show that Experiment I and II reach 91.27\% and 95.10\% accuracy scores respectively. It can be said that giving only the vehicles in a well-centered way to the model helps to increase accuracy significantly. It is interesting to see that Experiment II generalizes the Other Class 10\% better. Fiat Polo Class has less amount of images comparing to the other classes. However, Experiment II also generalizes it 11\% better than Experiment I. 

\begin{figure}[H]
\resizebox{\textwidth}{!}{%
\begin{subfigure}{0.5\textwidth}
\captionsetup{justification=centering}
\includegraphics[width=1\linewidth]{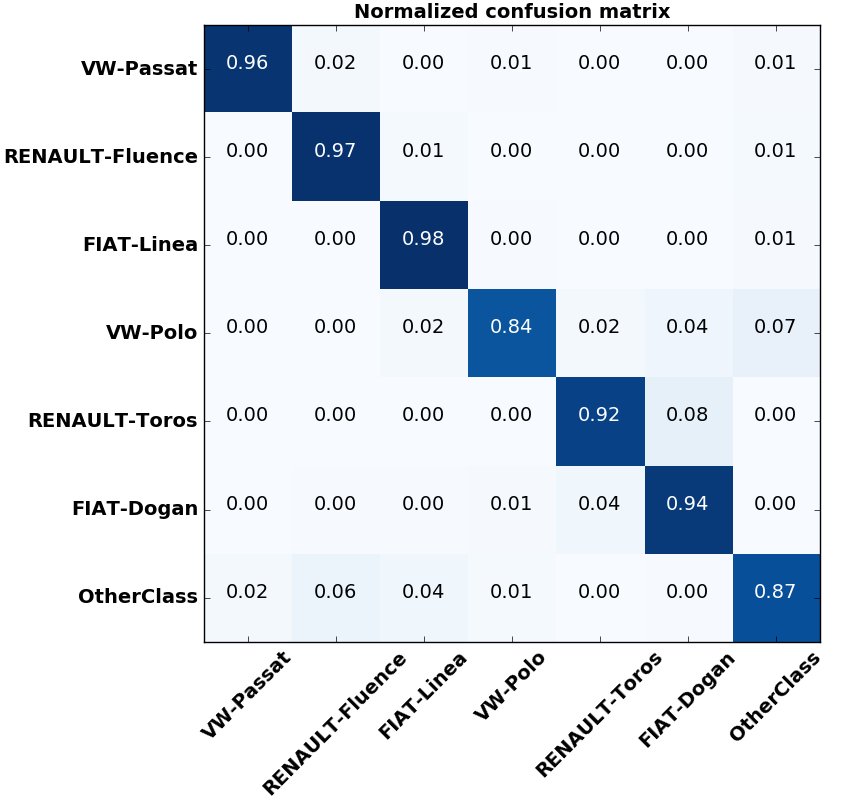} 
\caption{Only ResNet Model}
\label{fig:subim5}
\end{subfigure}
\begin{subfigure}{0.5\textwidth}
\captionsetup{justification=centering}
\includegraphics[width=1\linewidth]{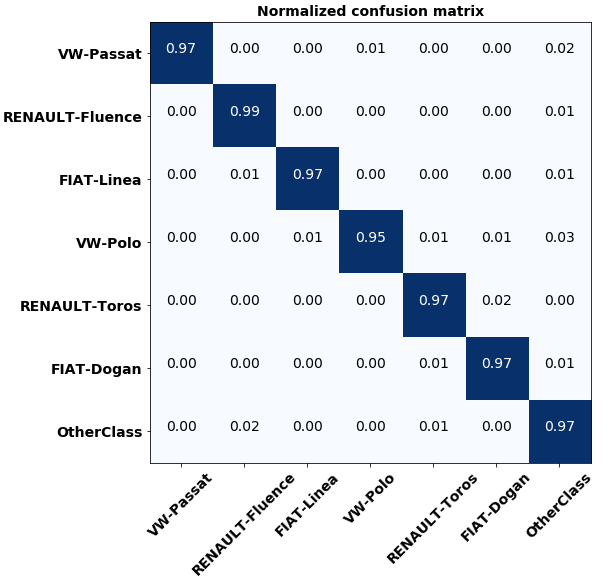}
\centering
\caption{Car Detection by Fine-tuned SSD + ResNet Model}
\label{fig:subim6}
\end{subfigure}
}
\captionsetup{justification=centering}
\caption{Confusion matrices: \\overall accuracy results of Experiment I and Experiment II}
\label{fig:image1}
\end{figure}

Less number of epochs are used in Experiment III than the others. The original SSD model has a VGG base which pre-trained on ImageNet first. Therefore, no need to train further. It is also seen that there is a significant difference between train and validation score. It indicates that we especially need more data and a bit of a change in our model.

Figure 5 shows that we reach 90.57\% mAP (Mean Average Precision) score on the test set. It is seen that Experiment III has almost the same and even better classification scores for several classes. We should note that GTBB of images is defined by an algorithm which uses a pre-trained SSD model. The Other Class has the lowest accuracy result. Finally, it reaches 70.34\% detection accuracy on the test set. Localization loss is also included in calculating this accuracy.

\begin{figure}[H]
\begin{center}
\resizebox{0.5\textwidth}{!}{

\includegraphics[width=0.7\linewidth]{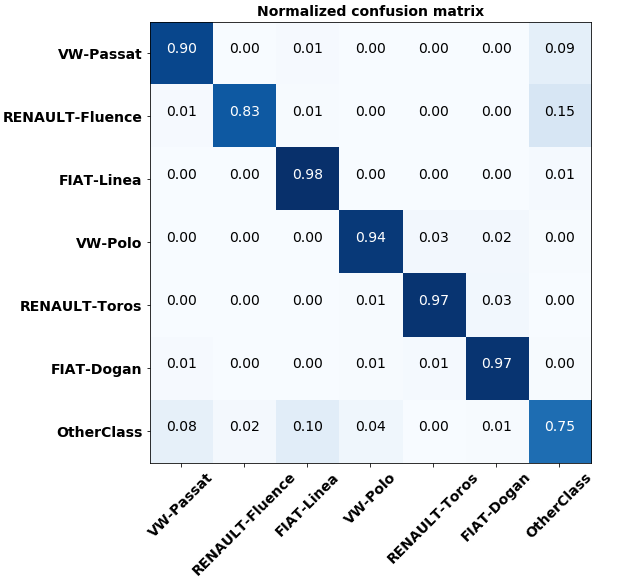}
}
\end{center}
\caption{Overall accuracy result of Experiment III, VGG based weights of SSD}
\end{figure}

We also tested the SSD based model on some videos. It can detect the vehicles with 12 FPS using Tesla K80. When we compare the power of GPUs with the one used in the original paper, it is very reasonable to get this FPS score. The results can be found in the author's repository. \url{https://goo.gl/EB6vyF}

Tables \ref{tab:trueDecision} and \ref{tab:falseDecision} show the certain positive and negative outcomes respectively. Green lines refer to the correct predictions. The first image shows that the model can predict the vehicle better in spite of not having perfectly shaped GTBB. The other samples show that having a well-centered position in an image have well effects on detection. 

Red lines refer to false predictions in Table \ref{tab:falseDecision}. The samples show that if the vehicles have a relatively small size in the image, it causes to be detected wrongly. Oppositely, having a relatively big size in the image also causes false detection. 

\begin{table}[]
\centering
\caption{True decisions}
\begin{tabular}{lcccc}
\hline
\multicolumn{1}{|l|} {Sample}
&   
\multicolumn{1}{c|}{
\begin{minipage}{0.193\textwidth}
    \includegraphics[width=\linewidth]{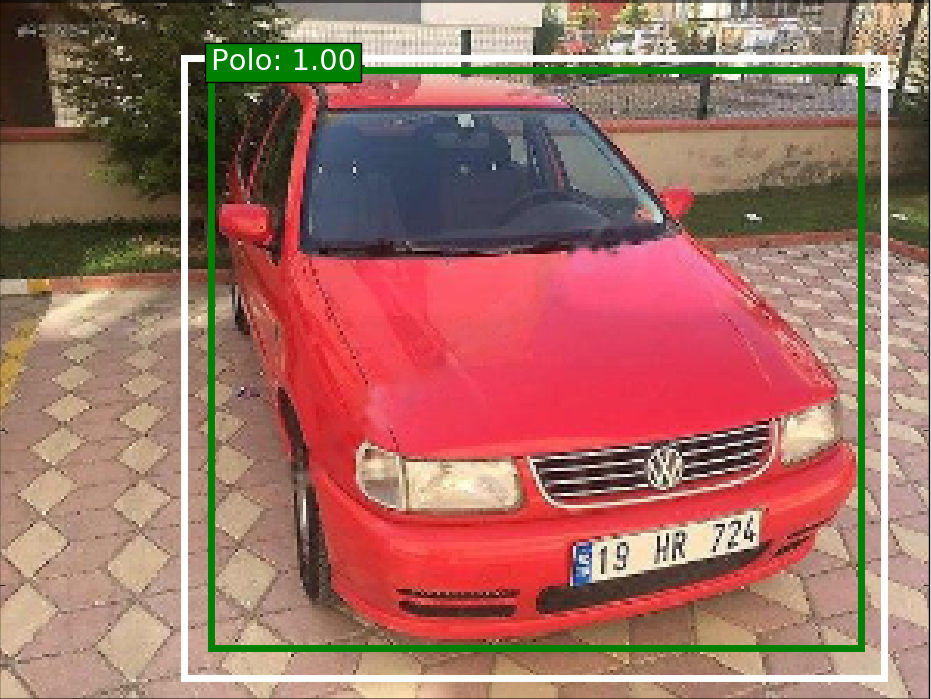}
\end{minipage} }
&
\multicolumn{1}{c|}{
\begin{minipage}{0.193\textwidth}
    \includegraphics[width=\linewidth]{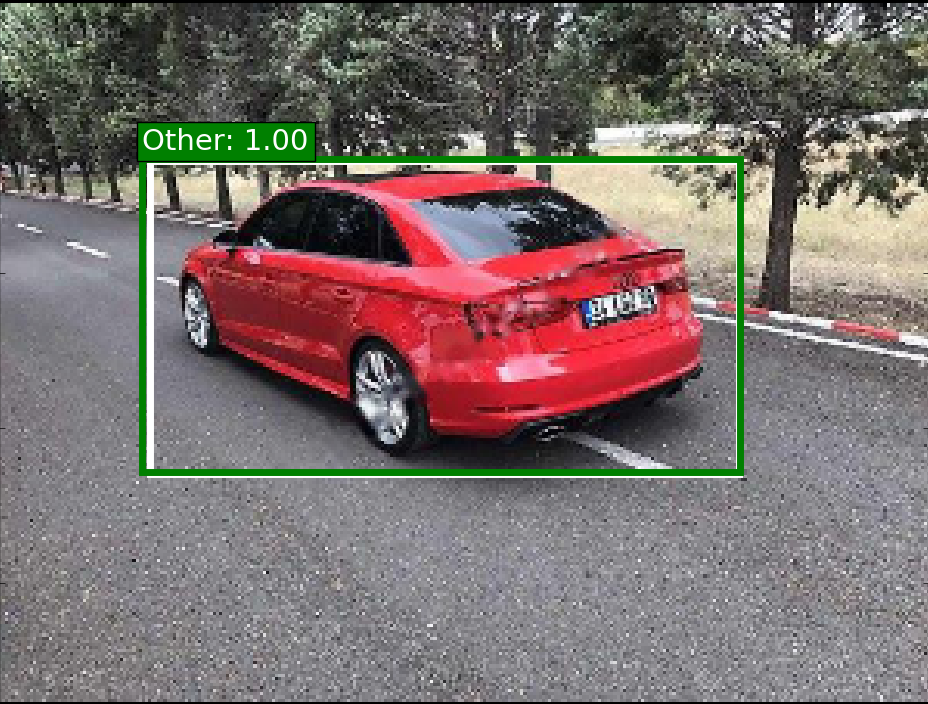}
\end{minipage}}
&
\multicolumn{1}{c|}{
\begin{minipage}{0.193\textwidth}
    \includegraphics[width=\linewidth]{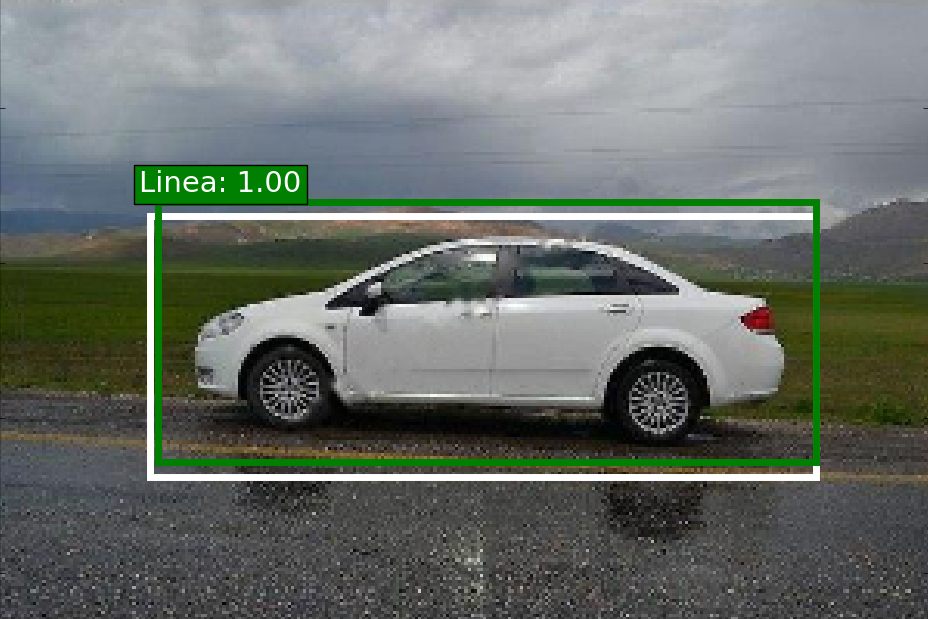}
\end{minipage}}
&
\multicolumn{1}{c|}{
\begin{minipage}{0.193\textwidth}
    \includegraphics[width=\linewidth]{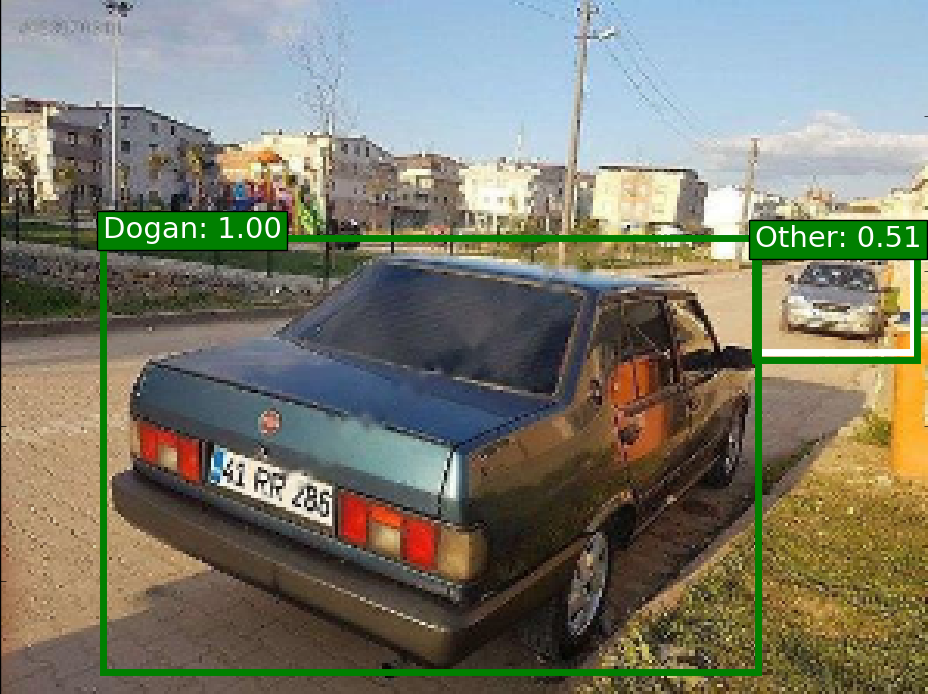}
\end{minipage}
}  \\ \hline
\multicolumn{1}{|l|}{Predicted Class}  & \multicolumn{1}{c|}{VW. Polo} & \multicolumn{1}{c|}{Other Class} & \multicolumn{1}{c|}{Fiat Linea} & \multicolumn{1}{c|}{Fiat Dogan} \\ \hline
\multicolumn{1}{|l|}{Probability} & \multicolumn{1}{c|}{1.00}  & \multicolumn{1}{c|}{1.00} & \multicolumn{1}{c|}{1.00} & \multicolumn{1}{c|}{1.00}  \\ \hline
& & \multicolumn{3}{r}{
\begin{minipage}{0.55\textwidth}
    \includegraphics[width=\linewidth]{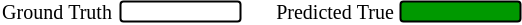}
\end{minipage}
}
\end{tabular}
\label{tab:trueDecision}
\end{table}

\begin{table}[]
\centering
\caption{False decisions}
\begin{tabular}{|l|c|c|c|c|}
\hline
Sample      
&   
\begin{minipage}{0.193\textwidth}
    \includegraphics[width=\linewidth]{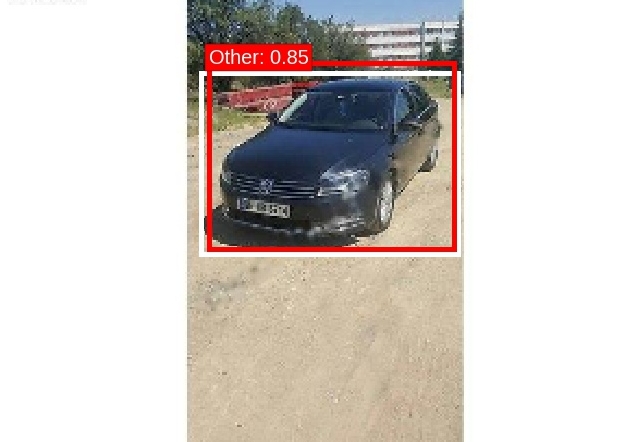}
\end{minipage}
&
\begin{minipage}{0.193\textwidth}
    \includegraphics[width=\linewidth]{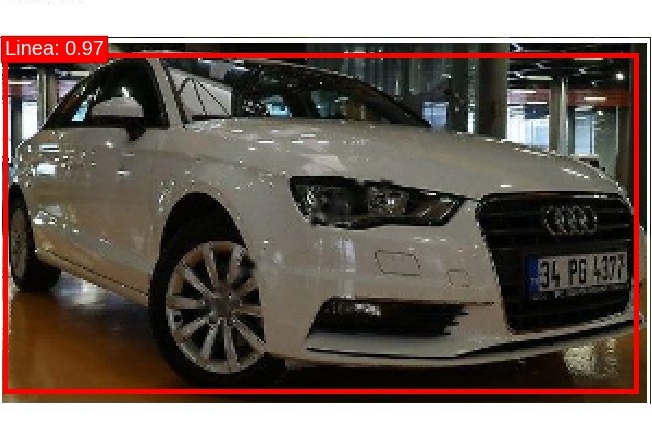}
\end{minipage}
&
\begin{minipage}{0.193\textwidth}
    \includegraphics[width=\linewidth]{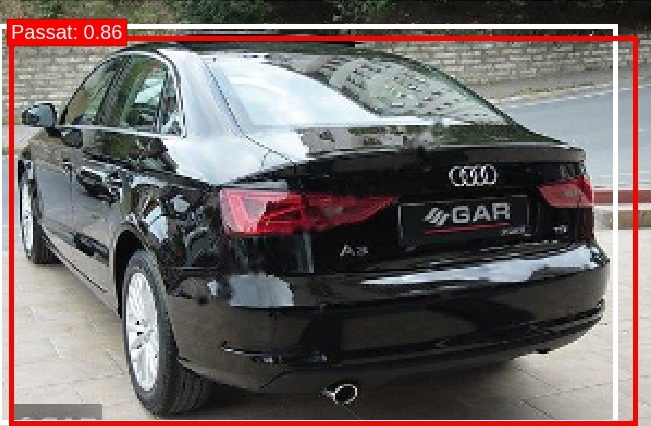}
\end{minipage}
&
\begin{minipage}{0.193\textwidth}
    \includegraphics[width=\linewidth]{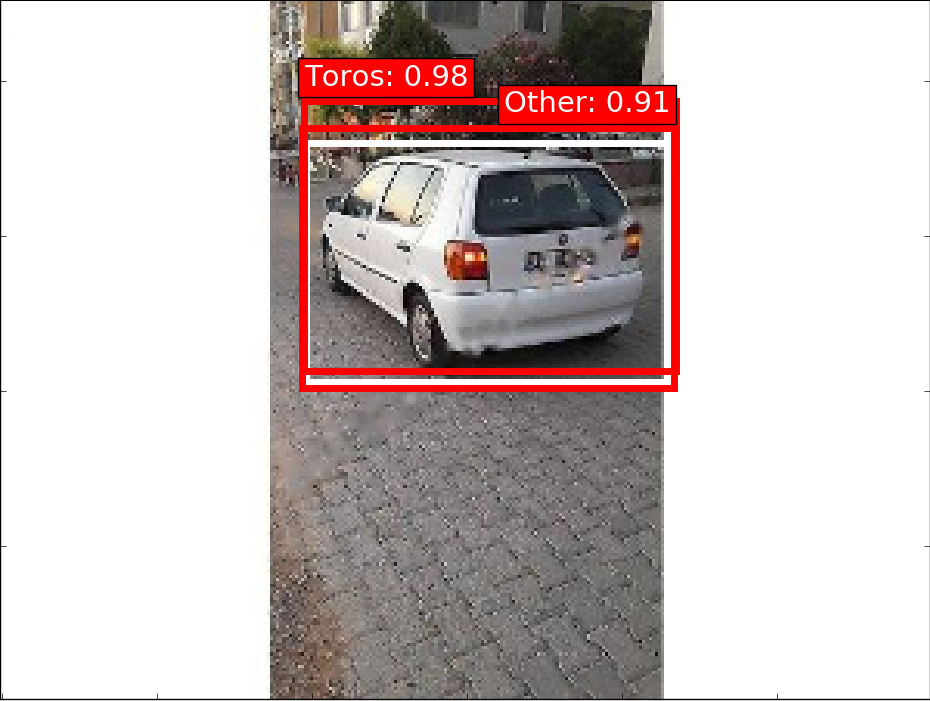}
\end{minipage}
\\ \hline
Correct Class    & VW. Passat & Other Class & Other Class & VW. Polo\\ \hline
Predicted Class   & Other Class & Fiat Linea & VW. Passat & Renault Toros \\ \hline
Probability    & 0.85    & 0.97     & 0.86     & 0.98       \\ \hline
\multicolumn{5}{r}{
\begin{minipage}{0.57\textwidth}
    \includegraphics[width=\linewidth]{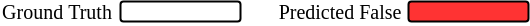}
\end{minipage}
}
\end{tabular}
\label{tab:falseDecision}
\end{table}

\subsection{A Use Case}

\begin{figure}[H]
\begin{center}
\includegraphics[width=0.55\linewidth]{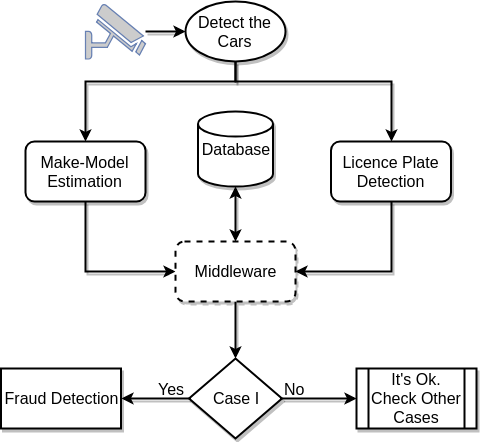}
\end{center}
\caption{A use case diagram, the case I: license plates match, but models don't }
\label{fig:useCase}
\end{figure}

This study also introduces an application for detecting fraudulent license plates in Figure \ref{fig:useCase}. Middleware first takes the detected license plate number and predicted class by using our make \& model detection method. Then it assigns them to a dictionary value which is composed of a key and value. Finally, it compares them with the values of the database.

It is assumed that every license plate number is matched with the class of a car and the database is set manually from the beginning. Then, if the license plate numbers are matched but make \& models of a car not, it means there is a fraud. Meaning that the license plates are recurrent and the detected car uses an illegal license plate. The performance of the application depends on how accurate and fast the detection and the prediction are made. 

\section{Conclusions and Future Works}

This study deals with the problems of vehicle make \& model classification. At first, a fine-grained database is created to gather a large number of samples specific to Turkey. It is used in the test experiments. As a first outcome of this paper, we see that combining a CNN based model with an SSD model could increase the classification score. 
This paper also introduces an algorithm to reduce the annotation time; however, it causes not to have perfectly shaped GTTB of the images. Nevertheless, we reach an acceptable classification score. For future work, it requires a change in the filter sizes of the SSD model architecture to fix false detection results.
Besides, this study shows that implementing this model on fraud detection of license plates is considerably possible in certain use cases. The application could be extended to further use cases for future works. For instance, it is quite hard to detect an illegal car when the license plates cannot be recognized to read. However, security providers easily improve their chance to catch the unapproved vehicles by knowing the make \& model of them.

\bibliographystyle{splncs04}
\bibliography{samplepaper}

\end{document}